\documentclass{article}
\usepackage{icml2008}

\usepackage{graphics}
\usepackage{epsfig}
\usepackage{float}

\usepackage{graphicx}
\usepackage{mlapa}

\icmltitlerunning{Conditional anomaly detection methods for patient--management 
alert systems}

\begin{document}
%

\twocolumn[
\icmltitle{Conditional anomaly detection methods \\ 
for patient-management 
alert systems}

\icmlauthor{}{} 
\icmladdress{{\bf Keywords}: anomaly detection,
             alert systems, monitoring, 
             health--care applications} 
%
 \icmlauthor{Michal Valko}{michal@cs.pitt.edu}
 \icmlauthor{Gregory Cooper}{gfc@cbmi.pitt.edu}
 \icmlauthor{Shyam Visweswaran}{shv3@pitt.edu}
 \icmlauthor{Milos Hauskrecht}{milos@cs.pitt.edu}
 \icmladdress{University of Pittsburgh, PA}

\vskip 0.3in
]

\newcommand{\A}{{\cal A}}
\newcommand{\coA}{\bar{A}}
\newcommand{\B}{{\cal B}}
\newcommand{\E}{{\cal E}}
\newcommand{\bX}{{\bf X}}
\newcommand{\bY}{{\bf Y}}
\newcommand{\bZ}{{\bf Z}}
\newcommand{\bW}{{\bf W}}
\newcommand{\bC}{{\bf C}}
\newcommand{\bB}{{\bf B}}
\newcommand{\bA}{{\bf A}}
\newcommand{\bx}{{\bf x}}
\newcommand{\ba}{{\bf a}}
\newcommand{\bb}{{\bf b}}
\newcommand{\bc}{{\bf c}}
\newcommand{\bv}{{\bf v}}
\newcommand{\bw}{{\bf w}}
\newcommand{\bu}{{\bf u}}
\newcommand{\by}{{\bf y}}
\newcommand{\bz}{{\bf z}}
\newcommand{\bq}{{\bf q}}
\newcommand{\bt}{{\bf t}}
\newcommand{\vecb}{{\bf b}}
\newcommand{\bT}{{\bf T}}

\newcommand{\mynote}[1]{\begin{center}\fbox{\parbox{5in}{#1}}\end{center}}
\newcommand{\commentout}[1]{}
\newcommand{\highl}[1]{\em {#1}}
\newcommand{\action}[0]{\eta}
\newcommand{\vect}[1]{\bf {#1}}
\newcommand{\real}{\hbox{\it I\hskip -2pt R}}



\commentout{
\begin{abstract}
Anomaly detection methods can be very useful in identifying unusual or interesting
patterns in data. A recently proposed conditional anomaly detection framework extends 
anomaly detection to the problem of identifying anomalous patterns on a 
subset of attributes in the data. The anomaly always depends (is conditioned) on the value of remaining 
attributes. The work presented in this paper focuses on instance--based methods
for detecting conditional anomalies. The methods depend heavily on the distance metric that lets us
identify examples in the dataset that are most critical for detecting the anomaly. To optimize the performance
of the anomaly detection methods we explore and study metric learning methods.  
We evaluate the quality of our methods on the Pneumonia PORT dataset by detecting unusual admission decisions
for patients with the community--acquired pneumonia. The results of
our metric learning methods show an improved detection performance over standard distance metrics, which is 
very promising for building automated anomaly detection systems for variety of intelligent monitoring
applications. 

\end{abstract}
}
\section{Introduction}

Anomaly detection methods can be very useful in identifying interesting or 
concerning events. Typical anomaly detection attempts to identify
unusual data instances that deviate from the majority of examples in the dataset. Such instances 
indicate anomalous (out of ordinary) circumstances, for example, a network attack \cite{eskin2000anomaly} or 
a disease outbreak \cite{wong2003bayesian}.  In this work, we study conditional anomaly detection  
framework that extends standard anomaly detection by identifying partial 
patterns in data instances that are anomalous with respect to the remaining data features. Such a framework 
has been successfully applied to identify unusual patient--management decisions made for patients
suffering from different conditions \cite{hauskrecht2007evidence-based}.  

Data attributes (features) in the conditional anomaly detection are divided into two disjoint groups: 
context (or condition) attributes $C$ and target attributes $A$. 
The conditional anomaly methods then attempt to identify anomalies in target attributes $A$ 
with respect to context $C$. The conditional aspect allows us to identify patterns that are
typical in one context but anomalous in the other. To illustrate the potential of the method assume two patients with different
conditions are given the same drug. In one of these conditions the administration of the drug is normal, but for the
other one it is unusual. The conditional anomaly detection methods with target variables corresponding to the treatment
should be able to identify the anomaly in the treatment. 

The conditional anomaly detection method evaluates and identifies anomalies 
one data example at the time. To make an anomaly 
call for a data instance $\bx$, \cite{hauskrecht2007evidence-based} proposed a probabilistic predictive model $M$ 
that aims to capture  stochastic dependencies among the target and context attributes. 
The predictive model defines a conditional probability distribution 
$p(A|C)$ of target attributes given the values of context variables. Given the predictive model, the 
anomaly call for a data instance $\bx$ is made if the probability of the target attributes 
observed in $\bx$ is small.   

A predictive probabilistic model used for detection purposes can be built in different ways. 
In this paper, we focus on instance--based approaches. The instance--based methods do not try to
learn a universal predictive model for all possible instances at the same time, instead the model
is optimized for every data instance $\bx$ individually. The instance--specific model $M_{\bx}$
may provide a better option if the predictive model is less complex and the dataset is small. 

Instance--specific models often rely on a distance metric that aims to pick examples most relevant for the prediction of $\bx$. 
However, the question of what is the best distance metric to reflect the relevancy of the example to the prediction 
is the most challenging part of the task.
Standard metrics such as Euclidean or Mahalanobis metrics are not the best for 
the anomaly detection task since they may be biased by feature duplicates or features that are irrelevant for predicting
target attributes. Thus, instead of choosing one of the standard distance metrics we  
investigate and test metric--learning methods that let us adapt  
predictive models to specifics of the currently evaluated example $\bx$. 

We explore two metric--learning methods that were originally used for building  
non--parametric classification models. The first method is NCA \cite{goldberger2004neighbourhood}. The method adjusts
the parameters of the generalized distance metric so that the accuracy of the associated nearest neighbor
classifier is optimized. The second method, RCA \cite{bar-hillel2005learning}, optimizes mutual information between the distribution in the original 
and the transformed space with restriction that distances between same class cases do not exceed a fixed threshold. 

To evaluate the quality of metric learning methods in anomaly detection we apply them to the problem of identification 
of unusual patient--management decisions, more specifically, to the problem of detection of unusual hospitalization patterns 
for patients with the community acquired pneumonia. We show that on this problem metric learning approaches 
outperform standard distance metrics. 

\commentout{
\section{Methodology}

\subsection{Conditional anomaly detection}

In anomaly detection, we are interested in detecting an unusual data pattern the occurrence of which 
deviates from patterns seen for other examples. In the conditional anomaly, a partial data pattern is evaluated
in context of other data variables and their values. Briefly, the data attributes (features)
are divided into two disjoint groups: context (or condition) attributes $C$ and the target attributes $A$. 
The objective of conditional anomaly detection methods is to identify anomalies in target attributes 
$A$ with respect to context attributes $C$.  

Let $E=\{\bx^1, \bx^2, \dots,  \bx^n\}$ be a set of examples in the dataset and let $\bx$ be an example we want to 
analyze and determine if it is conditionally anomalous with respect to examples in $E$.
The context of the example $\bx$ is defined by the projection 
of $\bx$ to context attributes $C$, which we denote by $C(\bx)$. Similarly, $A(\bx)$ denotes the projection of $\bx$ to target
attributes. 

Our goal is to identify the anomaly in $\bx$ with respect to examples in the dataset $E$. The examples and their relation to 
$\bx$ can be captured indirectly by an auxiliary probabilistic predictive model $M$. This approach was proposed   
recently by \cite{hauskrecht2007evidence-based}. The predictive model $M$ defines a conditional 
probability distribution of target variables given the value of context variables:
$p(A|C)$ and it is induced (learned) from examples in $E$. Given $M$ we say the case $\bx$ is {\em anomalous}
in target attributes $A$, if the probability $p(A(\bx) | C(\bx))$ for the model is small and falls 
below some threshold. In summary, stochastic relations in between the context and target
attributes observed in examples $E$ are incorporated into a probabilistic model $M$, which is in turn applied to example $\bx$. 
The anomaly is detected by evaluating the probability of target variable values for the $\bx$ 
given the values of its context variables $C(\bx)$ in model $M$.    

To build a working anomaly detection algorithm, we need to provide methods for building a probabilistic model $M$
from the dataset and methods for detecting the anomaly using the model.

\subsection{Building a probabilistic model}

Our conditional anomaly framework builds upon the existence of an underlying probabilistic model $M$  
that describes stochastic relations among context and target data attributes. We consider two types of models 
to achieve this task: (1) parametric and (2) non--parametric models. 

\subsubsection{Parametric predictive models.} 
In the parametric approach we assume a predictive model $p(A|C)$ is defined using a small set of parameters $\Theta$ 
that reflect accurately the stochastic relation among the context and target attributes expressed in data $E$. Examples of parametric models are: 
a Bayesian belief network \cite{pearl1988probabilistic}, a Naive Bayes model
\cite{domingos1997optimality}, Linear discriminant analysis 
\cite{hastie2001elements}or a 
logistic regression model \cite{hastie2001elements}. 
In this work we focus on the Naive Bayes model that is used frequently in classification tasks. We adopt 
the Bayesian framework to learn the parameters of the model from data $E$ and to support probabilistic inferences.  
In such a case the parameters $M$ of the model are treated as random variables and are described 
in terms of a density function $p(\theta_M)$. The probability of an event is obtained by 
averaging over all possible parameter settings of the model $M$. 
 
To incorporate the effect of examples $E$, $p(\theta_M)$  corresponds to the posterior $p(\theta_M |E)$. 
The posterior is obtained via Bayes theorem:
$$p(\theta_M | E) = p(E|\theta_M) p(\theta_M) / p(E),$$
where $p(\theta_M)$ defines the prior for parameters $\theta_M$. 
To simplify the calculations we assume \cite{heckerman1995tutorial} (1) parameter independence and (2) conjugate priors. 
In such a case, the posterior follows the same distribution as the prior and updating reduces to updates of 
sufficient statistics. Similarly, many probabilistic calculations can be performed in the closed form.

\subsubsection{Instance--specific models.}
In general, a predictive probabilistic model used for anomaly detection purposes can be of different complexity. 
However, if the dataset used to learn the model is relatively small, a more complex model may 
become hard to learn reliably. In such a case a simpler parametric model of $P(A|C)$ with a smaller number of 
parameters may be preferred. Unfortunately, a simpler model may sacrifice some flexibility and its predictions 
may become biased towards the population of examples that occurs with a higher prior probability.  
To make accurate predictions for any instance we use {\em instance--specific} predictive methods and models 
\cite{visweswaran2005instance-specific,aha1991instance-based}.

Briefly, instance--based methods do not try to
learn a universal predictive model for all possible instances, instead the model
is optimized for every data instance $\bx$ individually. To reflect this, we denote the predictive model for $\bx$ as $M_{\bx}$. 
The benefit of instance--based parametric models is that they can be fit more accurately to any data instance; the limitation is
that the models must be trained only on the data that are relevant for $\bx$. Choosing
the examples that are most relevant for training the instance--specific model is the 
bottleneck of the method. We discuss methods to achieve this later on.

\subsubsection{Non--parametric predictive models.} 

Non--parametric predictive models do not assume any compact parametrization of $P(A|C)$. 
Instead, the model is defined directly on the dataset of examples $E$.  A classic example of a non--parametric model
is the $k$ Nearest Neighbor ($k$--NN) classifier in which the predicted class of the instance is the majority vote of the 
classes of its $k$ nearest neighbors. 

Non--parametric models are instance specific by definition. For example, the $k$--NN classifier for instance $x$ 
executes by finding $k$ examples closest to $\bx$ first and making the prediction afterwards. The problem of finding 
the $k$ closest neighbors is the bottleneck of the method. Non--parametric models
depend on the choice of examples closest to $\bx$, and the quality of these choices influences the quality of the model. 

The anomaly detection approach applied in this work builds upon the model $M_{\bx}$ which defines the 
probability distribution $P(A|C(\bx))$ for $\bx$.
But how to define a non-parametric predictive model $M_{\bx}$? The key here is to define the probability 
with which a neighbor example predicts the values of target attributes A for $\bx$.  Intuitively, 
closer neighbors should contribute more and hence their prediction should come with a higher probability. 
To reflect this intuition \cite{goldberger2004neighbourhood} 
define the probability that a data example $\bx'$ predicts $\bf x$ using the softmax model \cite{mccullagh1989generalized}.
In this model, the probability with which $\bx'$ contributes to the prediction of $\bx$ is proportional to:
${\exp(-||\bx -  \bx'||_m^2)}$ where $m$ is a distance metric reflecting the similarity of the samples. 
The softmax model normalizes this quantities so that their sum is 1. 

The above definition of a non--parametric probabilistic predictive model expects a distance metric defining the similarity
among examples. We return to the problem of distance metrics in the next section.

\subsection{Anomaly detection}

Multiple approaches can be used to make anomaly calls based on the probabilistic metric. 
Typically, they rely on a variety of thresholds. These include: absolute, relative or the $k$ standard deviation thresholds. 
In our work, we build upon the absolute threshold test.  In the absolute threshold test, the example $\bx$ is anomalous 
if $p(A(\bx)|C(\bx),M_{\bx})$ falls below some  fixed probability threshold $p_\varepsilon$. Intuitively, if 
the probability of the target attributes $A(\bx)$ for $x$ is low with respect to the model $M_{\bx}$ 
and its other attributes $C(\bx)$, then the value of the target attribute is anomalous. Note 
that the absolute threshold test relies only on the model $M_{\bx}$ and there is no direct comparison 
of predictive statistics for $\bx$ and examples in $E$.  However, if instance--based methods are used the most important examples in $E$ 
are used to construct the model $M_{\bx}$ and hence their effect is reflected in the statistic. 

\section{Defining the similarity metric}

Parametric instance--based models are sensitive of examples used to train them. Similarly, the instance--based
non--parametric models are sensitive to examples incorporated into the model. The key question is what examples from $E$
should be used for training or defining the instance--specific predictive model $M_{\bf x}$.  
 
\subsubsection{Exact match.} Clearly, the best examples are the ones that exactly 
match the attributes $C(\bx)$, of the target case $\bx$.  However, it is very likely 
that in real--world databases none or only few past cases match the target case exactly so there is no 
or very weak population support to draw any statistically sound anomaly conclusion. 

\subsubsection{Similarity--based match.} 

One way to address the problem of insufficient population available 
through the exact match is to define a distance metric on the space of 
attributes $C(\bx)$ that let us select examples closest to the target example $\bx$. 
The distance metric defines the proximity of any two cases in the dataset, and 
the $k$ closest matches to the target case define the best population of size $k$. 
Different distance metrics are possible. An example is the generalized distance metric $r^2$ defined: 
\begin{equation}
r^2(\bx^i,\bx^j) = (\bx^i - \bx^j)^T\Gamma^{-1}(\bx^i - \bx^j), \label{eq:generalized_metric}
\end{equation}
where $\Gamma^{-1}$ is a matrix that weights attributes of patient cases proportionally to their importance.  
Different weights lead to a different distance metric. For example, if $\Gamma$   is the identity matrix $I$, 
the equation defines the Euclidean distance of $x^i$ relative to $x^j$. The Mahalanobis distance \cite{mahalanobis1936generalized} 
is obtained from \ref{eq:generalized_metric} by choosing  $\Gamma$ to be the population covariance matrix $\Sigma$  which lets us incorporate 
the dependencies among the attributes. 

The Euclidean and Mahalanobis metrics are standard off--shelf distance metrics often applied in many learning tasks. 
However they come with many deficiencies. The Euclidean metric ignores feature correlates
which leads to `double--counting' when defining the distance in between the points. The Mahalanobis distance resolves this
problem by reweighting the attributes according to their covariances. Nevertherless, the major deficiency of both 
Mahalanobis and  Euclidean metrics is that they may not properly determine the relevance of an attribute for predicting 
the target attributes. 

The relevance of context attributes for anomaly detection is determined by their influence on target attributes A. 
Intuitively, a context attribute is relevant for the predictive model if is able to predict changes in values of target 
attributes A. To incorporate the relevance aspect of the problem into the metric we adapt (learn) the parameters 
of the generalized distance metric with the help of examples in the dataset $E$.

\subsection{Metric--learning} 
The problem of distance metric learning in context of classification tasks has been studied 
by \cite{goldberger2004neighbourhood} and \cite{bar-hillel2005learning}. We adapt these metric learning methods
to support probabilistic anomaly detection. In the following we briefly summarize 
the two methods.  

\cite{goldberger2004neighbourhood} explores the learning of the metric in context of the nearest neighbor
classification. They learn a generalized metric:
\begin{eqnarray*}
d^2(x_1,x_2) & = & (x_1 -x_2)^TQ(x_1-x_2) \\ & = & (x_1 -x_2)^TA^TA(x_1-x_2) \\
& =  & (Ax_1 -Ax_2)^T(Ax_1-Ax_2)\\
\end{eqnarray*}
by directly learning its corresponding linear transformation $A$. They introduce a new optimization criterion (NCA),
that is, as argued by the authors, more suitable for the nearest--neighbor classification purposes. 
The criterion is based on a new, probabilistic version of the cost function for the leave--one--out 
classification error in the $k$--NN framework. Each point $i$ can now select any other point $j$ with some probability $p_{ij}$ defined as
softmax function over distances in the transformed space:
$$p_{ij} = \frac{\exp(-||Ax_i -  Ax_j||^2)}{\sum_{k\ne i}\exp((-||Ax_k - Ax_j||^2)}$$
A linear transformation $A$ is then sought to maximize the expected number of
correctly classified cases (with $k$--NN): 
$$\arg\max_A g(A) = \arg\max_A \sum_i\sum_{j\in C_i} p_{ij}$$ 
where $C_i$ is the set of cases that belong to the same class as $i$. 
Intuitively, the criterion aims to learn a generalized distance metric by 
shrinking the distance between similar points to zero, and expanding the 
distance between dissimilar points to infinity.

The algorithm  and the metric it generates was shown to outperform other metrics for a number of 
learning problems. The method climbs the gradient of $g(A)$, which is ($x_{ij}$ being $x_i - x_j$):
$$\frac{\partial g}{\partial A} = 2A \sum_i\left(p_i\sum_k p_{ik}x_{ik}x_{ik}^T-\sum_{j\in C_i}p_{ij}x_{ij}x_{ij}^T\right)$$

\cite{bar-hillel2005learning} and \cite{shental2002adjustment} 
define a different optimization criterion based on the mutual information. 
The advantage of their method (relevant component analysis -- RCA) is the existence of the 
closed form (efficient) solution. Briefly, under the mutual information criterion, the class information 
is incorporated and optimized by computing the averages of class covariance matrices. The resulting matrix is obtained by 
\begin{equation}
\Sigma_{\mathrm{RCA}} = \sum_{i=1}^{k} \hat \Sigma_i \qquad A = \Sigma^{-\frac12} 
\end{equation}
where $\hat \Sigma_i$ sample covariance matrix of class $i$ and $A$ is the resulting transformation for the data. 
The disadvantage of the method is that it assumes Gaussian distribution for the classes. 

}

\section{Experimental evaluation}

We study our metric--learning methods and compare them to alternative methods on the problem of identification 
of anomalous patient--management decisions for patients with community acquired pneumonia.  
The data used in the experiment come from the Pneumonia PORT dataset \cite{kapoor1996assessment,fine1997prediction}. 
The Pneumonia PORT dataset is based on the study conducted from October 1991 to March 1994 on 2287 patients with community--acquired 
pneumonia from three geographical locations at five medical institutions. The original PORT data were 
analyzed by \cite{fine1997prediction}, who derived a prediction rule with $30$--day hospital 
mortality rate as the outcome. The authors developed a logistic regression model, which helped to identify 
20 attributes that contribute the most to the mortality rate of pneumonia. To explore the anomaly
detection methods, we have experimented with a simpler version of the PORT dataset that records, for every patient, only 
the attributes identified by Fine's study \cite{fine1997prediction}. 
\commentout{
The attributes are 
summarized in Figure \ref{fig:PORT-attributes}. All attributes are binary with true / false (positive / negative) values. 
}
Our objective was to detect unusual admission decisions (treat the patient at home versus in the hospital).
\commentout{
which are captured by the variable 'Hospitalization'.
}

\subsection{Study design}

To study the performance of our anomaly detection methods, we used 100 patient cases (out of a total of 2287 of cases). 
The cases picked for the study consisted of 21 cases that were found anomalous according to a simple Naive Bayes detector (with 
detection threshold 0.05) that was trained on all cases in the database. The remaining 79 cases were selected randomly from the rest of the database.  Each of the 100 cases was then evaluated 
independently by a panel of three physicians. The physicians were asked whether they agree with the hospitalization decision or not. 
Using panel's answers, the admission decision was labeled as anomalous when (1) at least two physicians disagreed with the actual 
admission decision that was taken for a given patient case or (2) all three indicated they were unsure (gray area) 
about the appropriateness of the management decision. Out of 100 cases, the panel judged 23 as anomalous hospitalization d
ecisions; 77 patient cases were labeled as not being anomalous.  The assessment of 100 cases by the panel represented 
the correct assessment of unusual hospitalization decisions.

\commentout{
\begin{figure}
  \centering
  {\small
  \begin{tabular}{l l} \hline
    \multicolumn{2}{c}{\textbf{Target attributes}} \\ \hline
    & \\
    $X_{1}$ & Hospitalization \\
    & \\ \hline
    \multicolumn{2}{c}{\textbf{Prediction attributes}} \\ \hline
    & \\
    & \textbf{Demographic factors} \\
    $X_{2}$ & \hspace{0.1in}Age $>$ 50 \\
    $X_{3}$ & \hspace{0.1in}Gender (male = true, female = false) \\
    & \textbf{Coexisting illnesses} \\
    $X_{4}$ & \hspace{0.1in}Congestive heart failure \\
    $X_{5}$ & \hspace{0.1in}Cerebrovascular disease \\
    $X_{6}$ & \hspace{0.1in}Neoplastic disease \\
    $X_{7}$ & \hspace{0.1in}Renal disease \\
    $X_{8}$ & \hspace{0.1in}Liver disease \\
    & \textbf{Physical-examination findings} \\
    $X_{9}$ & \hspace{0.1in}Pulse $\geq$ 125 / min \\
    $X_{10}$ & \hspace{0.1in}Respiratory rate $\geq$ 30 / min \\
    $X_{11}$ & \hspace{0.1in}Systolic blood pressure $<$ 90 mm Hg \\
    $X_{12}$ & \hspace{0.1in}Temperature $< 35\,^{\circ}$C or $\geq 40\,^{\circ}$C \\
    & \textbf{Laboratory and radiographic findings} \\
    $X_{13}$ & \hspace{0.1in}Blood urea nitrogen $\geq$ 30 mg / dl \\
    $X_{14}$ & \hspace{0.1in}Glucose $\geq$ 250 mg / dl \\
    $X_{15}$ & \hspace{0.1in}Hematocrit $<$ 30\% \\
    $X_{16}$ & \hspace{0.1in}Sodium $<$ 130 mmol / l \\
    $X_{17}$ & \hspace{0.1in}Partial pressure of arterial oxygen $<$ 60 mm Hg \\
    $X_{18}$ & \hspace{0.1in}Arterial pH $<$ 7.35 \\
    $X_{19}$ & \hspace{0.1in}Pleural effusion \\
    & \\ \hline
  \end{tabular}
  }
  \caption{Attributes from the Pneumonia PORT dataset used in the
  anomaly detection study.}
  \label{fig:PORT-attributes}
\end{figure}
 }

\subsection{Experiments} 
All the experiments followed the leave--one--out scheme. That is, for each example in the dataset 
of 100 patient cases evaluated by the panel, we first learn the metric. Next, we identified 
the cases in $E$ most similar to it with respect to that metric. 
The cases chosen were either the closest $40$ cases, or all the other cases ($2286$) in the dataset. 
We then learned the NB model and calculated the posterior probability of the decision. 
Alternatively, we calculated the probability of the decision using the softmax model and  
the learned metric. 

The target example was declared anomalous if its posterior probability 
value fell below the detection threshold.  The anomaly calls made by our algorithms 
were compared to the assessment of the panel and resulting statistics (sensitivity, specificity) 
were calculated. To gain insight on the overall performance of each method we varied 
its detection threshold and calculated corresponding receiver operating characteristic (ROC). 
For the hospital deployment no all thresholds are acceptable. Consequently, for the evaluation 
we selected only that part of the ROC curve, that corresponds to specificity equal 
or greater than 95\% (see Figure \ref{fig:roc}). The 95\% specificity limit means that at most 1 in 20 normal cases analyzed may yield a false alarm. 

\section{Results and Discussion}
  \begin{table}

\begin{tabular}{|r|r|r|r|}
\hline

\hline
{\bf model }	   &    {\bf selection}        &   {\bf}         &  {\bf}    \\
\hline
\hline
{\bf non--parametric }	   &    {\bf global}        &   {\bf \#cases}         &  {\bf area}    \\
\hline
      NCA  &    softmax &       2286 &     18.0 \% \\

     Mahal &    softmax &       2286 &     12.2 \% \\

       RCA &    softmax &       2286 &     11.6 \% \\

    Euclidean &    softmax &       2286 &     8.0 \% \\
\hline
\hline
     {\bf non--parametric } &   {\bf local}       &    {\bf \#cases}        &     {\bf area}         \\
\hline
      NCA  &    softmax &         40 &     20.2 \% \\

     Mahalanobis &    softmax &         40 &     15.0 \% \\

       RCA &    softmax &         40 &     12.8 \% \\

    Euclidean &    softmax &         40 &     8.0 \% \\
\hline
\hline
  {\bf parametric } &     {\bf global} &   {\bf \#cases}         &     {\bf area}         \\
\hline
       any  &         NB  &       2286 &     11.6 \% \\
\hline
\hline
  {\bf parametric } &     {\bf local} &    {\bf \#cases}        &     {\bf area}         \\
\hline
      NCA  &         NB &         40 &     16.8 \% \\

     Mahalanobis &         NB &         40 &     17.6 \% \\

       RCA &         NB &         40 &     17.6 \% \\

    Euclidean &         NB &         40 &     16.4 \% \\
\hline

\end{tabular}  

    \caption{Area under the ROC curve in the feasible range of 95\% -- 100\% specificity. Please note that the baseline 
 value for the random choice is 2.5\%. }	
     \label{tbl:res}
  \end{table}

Table \ref{tbl:res} shows the ROC statistics for the feasible detection range. 
For the softmax model, the NCA metric outperformed all other methods, whether it 
was using all cases (patients) or just the closest 40. 
We ascribe it to the fact, that NCA uses class information to weigh the features.  
The only other method that used class information was RCA. 
However, RCA uses class information only to consider (and average) covariance matrices for each class separately. Therefore, 
it still treats all features within the class the same way as the Mahalanobis metric, assuming the 
same relevance of all features. Comparing the global (all other patients) and local (closest 40 patients), 
local did always better: Close patients let us fit better the predictive model to the target patient, 
while taking all samples into the consideration biases the population.  The local methods were also better 
for the Naive Bayes model. They were also more robust with respect to the metric. The intuition 
behind this result is that when using NB model, all cases are treated the same way, the metric 
was only used to select them. On the other hand, in softmax model, the distance from the case 
in hand does matter and the method treats closer patients with a higher weight. Accordingly, 
it is more sensitive to the metric changes. 

\commentout{
\begin{figure}
 \begin{center}
      \includegraphics[width=8cm, clip,viewport=96 235 511 565]{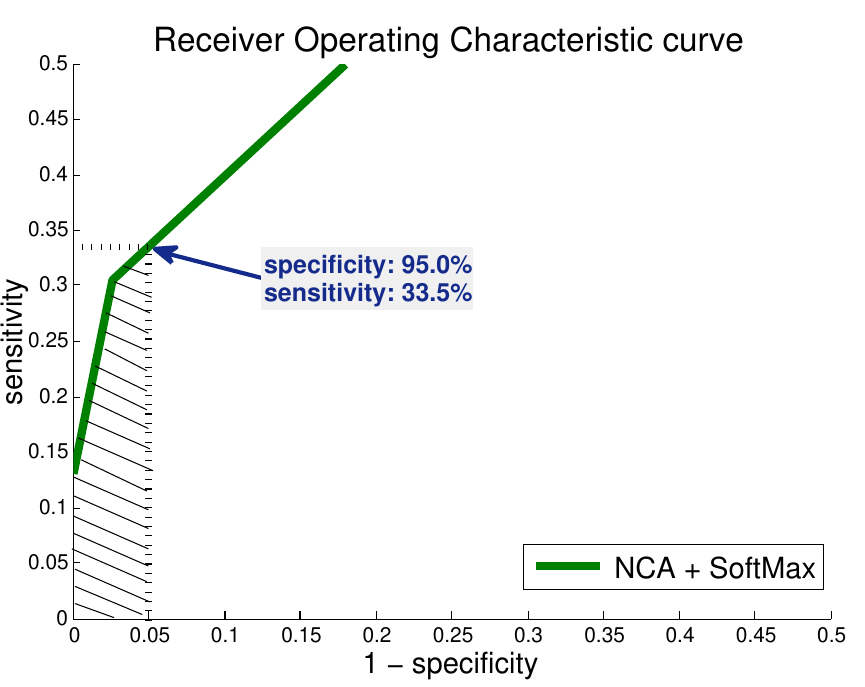}
      \caption{An example of the ROC curve for the method that performed the best on the pneumonia dataset. 
The statistic of interest is the leftmost region of the ROC curve and its area. }
    \label{fig:roc}
 \end{center}
 \end{figure}

Figure \ref{fig:roc} shows the ROC curve for the best method in Table \ref{tbl:res}. 
The area of interest is bounded by the values [0, 0.13], [0.03, 0.30], and [0.05, 0.33].
The point [0.05 0.33] corresponds to the performance of 6.66 correct alarms in 10 alarms
for 100 evaluated patients. However, we note that the prior for the evaluation dataset was biased towards anomalies. A 
rough correction using only anomalies that were randomly selected from the full database yields approximately 1 correct in 4
alarms, which is still very encouraging performance. 
}

\section{Conclusions}

Summing up, our conditional anomaly detection is a very promising methodology 
for detecting unusual events such as network attacks or medical errors. We have demonstrated
its potential by exploring and analyzing patient--management decisions for a dataset of patients  
suffering from pneumonia. The advantage of the anomaly detection approach over knowledge--based
error detection approaches is that the method is evidence--based, and hence requires no or minimum input 
from the domain expert.

Despite initial encouraging results, our current approach can be further refined and extended. 
For example, instance--based (local) models tested in this
paper always used a fixed number of 40 closest patients (or more, if the distances were the same). 
However, the patient's \emph{neighborhood} and its size depend on the patient and data available in the database.  
We plan to address the problem by developing methods that are able to automatically identify and select
only patients that are close enough for the case in hand. 

\section{Acknowledgements}

The research presented in this paper was funded by the grant R21--LM009102--01A1 from the National Library of
Medicine.  


\bibliography{miki}
\bibliographystyle{mlapa}

\end{document}